\documentclass[wcp]{jmlr}

\usepackage{booktabs}
\usepackage{array}

\makeatletter
\let \@jmlrpages \@empty
\makeatother

\usepackage{xspace}
\newcommand{\method}{TG-Mem\xspace}

\jmlrvolume{353,}
\jmlryear{2026}
\jmlrworkshop{ACML 2026}

\title[When Does Experience Memory Help LLM Driving Agents?]{When Does Experience Memory Help\\ Closed-Loop LLM Driving Agents?\\ {\large A Controlled Evaluation with Resampled Warmup Pools,\\ Token-Matched Fillers, and Audited Baselines}}

\author{\Name{Yingxin Jia} \Email{jia\_yingxin@hotmail.com}\\
\Name{Xuanyi Li} \Email{lixuanyi199801@gmail.com}\\
\addr Independent Researchers}

\editors{Andy Song, Bo Han and Sarah Erfani}

\begin{document}

\makeatletter
\let \@jmlrpages \@empty
\makeatother

\maketitle

\begin{abstract}
When does experience memory help a closed-loop LLM driving agent? Published gains cannot say, because planners, prompts, and implementation quality vary together
across systems, and the two factors we find to matter most, \emph{which} warmup experiences the pool
contains and what \emph{any} injected tokens do to the prompt, are controlled in none of them. We answer with a controlled evaluation on one
audited \texttt{HighwayEnv} testbed (kinematic observations, discrete meta-actions, $1$\,Hz,
\texttt{gpt-4o-mini} for the decisive contrasts). We freeze the planner, encoder, and protocol, treat
the warmup pool as a resampled \emph{random factor} (ten disjoint pools), and add token-matched
\emph{irrelevant-filler} controls, oracle-certified diagnostics, scripted certificate policies,
caution-pricing metrics, and quantified audits of the baseline ports. The probe suite spans flat-text retrieval (a DiLu port), a single-frame grid port
(RESPOND-inspired), our episodic temporal-graph memory \method{} with five variants, a retrieval-free
consolidated rulebook, and a flat $K$-step observation-history baseline. Four findings follow, each a
property of this testbed. \textbf{(1)} Headline memory gains belong to the warmup draw. The reported pool is the best of
ten for both retrieval memories, which beat No-Memory in only one pool of ten (sign test
$p\!=\!0.011$) and have a negative pool-averaged effect, and even the rulebook's apparent pool-immunity
($0.97$ to $0.98$ on five pools) collapsed to $0.03$ to $0.98$ on five fresh draws. \textbf{(2)} No
retrieval design tested exceeds token-matched filler, which sits significantly above both memories under
paired tests. The collision-only merge ceiling is reached by a scripted always-\texttt{SLOWER} policy and
by a single $15$-token caution instruction (both $0.98$), and metrics that price caution erase the
rulebook's lead. \textbf{(3)} Representational sophistication is nowhere robustly operative. Neither
retrieved structure nor raw in-context history solves the oracle-solvable history-necessity diagnostic.
\textbf{(4)} Baseline port quality silently inverts conclusions. On this testbed, a single-pool
comparison without filler controls would have reported memory gains that these controls reverse. We
release the testbed, diagnostics with certificates, the replay benchmark, audited ports, seeds, and caches.

\end{abstract}
\begin{keywords}
Large language models; autonomous driving; agent memory; temporal knowledge graphs; retrieval-augmented decision making; closed-loop simulation.
\end{keywords}

\section{Introduction}
\label{sec:intro}

\begin{figure}[t]
\begin{center}
\includegraphics[width=0.96\textwidth]{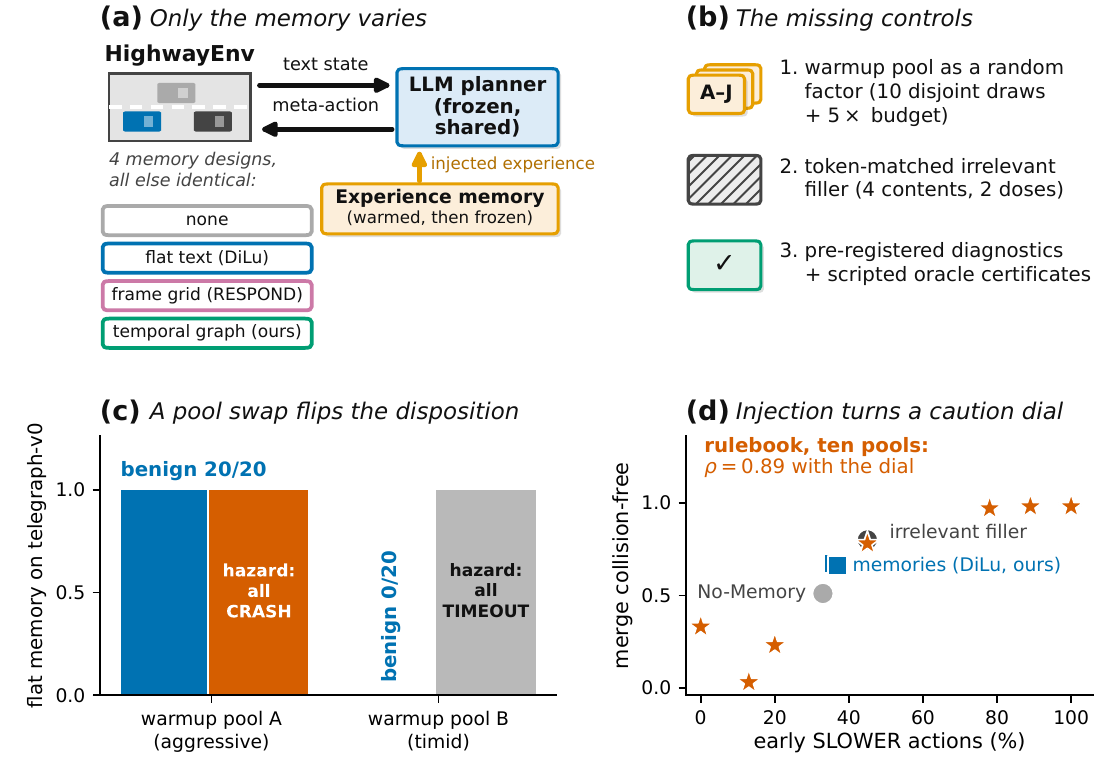}
\caption{Overview. \textbf{(a)} Memory designs from our probe suite and from the literature drive one
frozen planner in closed loop, and only the memory system varies. \textbf{(b)} The controls we
introduce: the warmup pool resampled as a random factor, token-matched irrelevant filler, and
pre-registered diagnostics with scripted oracle certificates. \textbf{(c)} Swapping one $8$-episode
warmup pool flips the flat memory from aggressive (all hazard episodes crash) to timid (all time out) on
the history-necessity diagnostic. \textbf{(d)} The early-\texttt{SLOWER} share tracks merge safety. The
retrieval-free rulebook turns this dial hardest, yet five fresh pools swing it from $0.03$ to $0.98$.}
\label{fig:overview}
\end{center}
\end{figure}

\looseness=-1 Large language models (LLMs) have emerged as promising decision makers for autonomous driving that
reason over scenes in natural language. A recurring finding is that they improve substantially when given
an \emph{experience memory}, a store of past situations and the actions that worked, retrieved and injected
as few-shot guidance at decision time~\citep{wen2024dilu,fu2024limsimpp,mei2024leapad}. This paper asks the question those reports leave open. \emph{When} does experience memory help,
and under which warmup experiences, prompt compositions, and decision cadences? Published comparisons
cannot answer this, because the memory varies together with the surrounding architecture and with
implementation quality, and the two factors we find to matter most, \emph{which} experiences the warmup
pool contains and what \emph{any} injected tokens do to the prompt, are controlled in none of them.

Our primary contribution is the \textbf{controlled evaluation} itself. We fix one planner, one encoder,
and one warm-up, freeze, and evaluate protocol, and we instrument it with controls this literature has
not used. The warmup pool becomes a \emph{resampled random factor} (ten disjoint pools with model-free
pool-level tests). \emph{Token-matched irrelevant-filler} arms span content, dose, and a
disclaimer$\times$domain factorial. Two constructed diagnostics are certified well-posed by scripted
oracles \emph{before} any LLM run. Scripted non-LLM certificate policies bound what caution alone
achieves, metrics price caution rather than only collisions, and a quantified audit measures the
fidelity of the baseline ports. On this testbed we evaluate a \textbf{probe suite} that spans the memory
design space (\S\ref{sec:method} and \S\ref{sec:setup}). At the rich end sits \method{} (Temporal-Graph
Memory), our episodic design. Published driving memories retrieve flat text by embedding
cosine~\citep{wen2024dilu,fu2024limsimpp,safedrive2024,mei2024leapad} or a single-frame
grid~\citep{respond2025}, whereas \method stores \emph{episodes} as small typed temporal graphs matched
by a combined structural and temporal score, a prompt-level driving instantiation of the temporal
knowledge-graph memory line~\citep{zep2025,mem0_2025,axu2025amem}, with five variants that toggle
curation, keys, and injection. At the distilled end sits a \textbf{consolidated rulebook}: the same
experience compressed \emph{once} into at most eight rules~\citep{zhao2024expel} and injected at every
decision with no retrieval. At the direct end sits a \textbf{flat $K$-step history} baseline, and at the
content-free end a \textbf{token-matched padding} arm and a four-content \textbf{filler suite}. Each
probe is both a design under evaluation and a control for its neighbors.

The evaluation yields four findings (Figure~\ref{fig:overview}). Each is a statement about this
testbed, \texttt{HighwayEnv} with kinematic observations, discrete meta-actions, a $1$\,Hz decision
cadence, and \texttt{gpt-4o-mini} for every decisive contrast, and each baseline statement concerns our
audited ports under this harness (\S\ref{sec:setup}), not the published systems. Table~\ref{tab:evidence}
maps each finding to the confirmatory evidence behind it.

\textbf{(M1) The warmup pool acts as a lottery: the headline gains are knife-edges that resampling
reverses.} Both headline gains, the diagnostic's $0.45\!\to\!0.95$ ($p{=}0.001$) and the merge margin of
$+0.15$ (significant under pairing), are properties of a particular draw. Resampled, the diagnostic
effect spans $0.95$ to $0.10$, and on merge the reported pool is the best of ten disjoint pools: each
memory exceeds No-Memory in one pool of ten (exact sign test $p\!=\!0.011$), sits significantly below it
in five, and has a pool-averaged effect near $-0.20$ (\S\ref{sec:results}). A pool swap also flips the
flat memory's entire \emph{disposition} from aggressive to timid on the history-necessity diagnostic
(Figure~\ref{fig:overview}c), and the rulebook's ``pool-immunity,'' confirmed on five pools, reversed on
five fresh draws.
\textbf{(M2) No retrieval design tested exceeds our token-matched fillers. Injecting content-free
tokens produces merge-safety changes comparable to the memories'.} Fillers span $0.59$ to $0.80$ on merge against the memories' $0.66$ to $0.67$, and the strongest
filler is significantly above both under paired tests. The designs that reach the collision-only
ceiling turn the caution dial hardest, and the dial needs no memory: a scripted always-\texttt{SLOWER}
policy and a single $15$-token caution instruction both score $0.98$, and the rulebook's best cells sit
at that operating point while metrics that price caution erase its lead (\S\ref{sec:consolidated}).
\textbf{(M3) The operative ingredient dissociates (curation on merges, key precision on the
diagnostic), and representational sophistication is nowhere robustly operative.} \method{} and its five
variants never robustly beat flat memory across splits. Its temporal edges, the design's central
component, contribute at most marginally in a pool-conditional ablation. The key$\times$injection
$2\!\times\!2$ (Table~S44) finds compact injection consistently harmful while the key-type attribution
flips with the pool, and even raw temporal context handed to the planner directly does not pass the
history-necessity diagnostic that memory was hypothesized to solve.
\textbf{(M4) Baseline implementation quality silently inverts conclusions.} A subtly crippled port
halved the flat baseline and manufactured a false win for our own method. We quantify the artifact and
release the audited ports.

\begin{table}[t]
\centering
\footnotesize
\setlength{\tabcolsep}{4pt}
\caption{Evidence map: each finding, the confirmatory result behind it, and the test used
($n\!=\!100$ paired seeds on merge, $n\!=\!40$ on diagnostics; $n\!\leq\!20$ cells are exploratory).}
\label{tab:evidence}
\begin{tabular}{@{}>{\raggedright\arraybackslash}p{0.08\textwidth}>{\raggedright\arraybackslash}p{0.54\textwidth}>{\raggedright\arraybackslash}p{0.34\textwidth}@{}}
\toprule
Finding & Confirmatory evidence & Test; location \\
\midrule
M1 & Ten disjoint $8$-episode pools, $n\!=\!100$ each: DiLu port and \method{} above No-Memory in $1/10$ pools, below in $5/10$; pool-averaged effect $-0.20$ [$-0.29,-0.10$] / [$-0.30,-0.09$]. Rulebook $0.97$--$0.98$ on A--E, $0.03$--$0.98$ on pre-registered F--J. & Exact sign test ($p\!=\!0.011$); per-pool tests, Holm; bootstrap over pools; Fig.~\ref{fig:pools}, Tab.~S45, S51, S52 \\
M2 & Padding filler $0.800$ vs.\ \method{} $0.660$ / DiLu port $0.670$ ($n\!=\!100$ paired); caution line $0.980$ above every memory and filler; always-\texttt{SLOWER} certificate $0.980$ (no LLM); rulebook SaS $0.00$. & McNemar ($p\!=\!0.029$, $0.024$; $\leq\!4\times10^{-9}$); Tab.~\ref{tab:main}, S47, S50 \\
M3 & DiLu port vs.\ \method{} indistinguishable ($n\!=\!100$ paired); every arm $0/20$ hazard successes on the oracle-solvable \texttt{telegraph-v0} ($n\!=\!40$, pre-registered); $-$temporal edges $0.55$ vs.\ $0.66$ ($n\!=\!40$, pool A). & McNemar ($p\!=\!1.0$; $0.07$); Tab.~\ref{tab:main}, S30 \\
M4 & Crippled DiLu port $0.375$ vs.\ audited port $0.670$; grid port fails two of three reproduction audits ($51\%$ decisions, $49\%$ cells reproduced). & Port audits; Tab.~S11 \\
\bottomrule
\end{tabular}
\end{table}

\paragraph{Contributions.}
\textbf{(i) An evaluation framework} for agent experience memory, with instruments this literature
lacks: the warmup pool as a resampled random factor with model-free pool-level tests, token-matched
filler and padding arms with a disclaimer$\times$domain factorial (the nearest analog is exemplar
sensitivity in in-context learning~\citep{lu2022fantastically}), oracle-certified diagnostics, scripted
certificate policies, caution-pricing metrics, and a quantified baseline audit.
\textbf{(ii) A probe suite} spanning the design space, all ours: \method, the first episodic
temporal-graph memory for closed-loop driving, with five variants, plus the consolidated rulebook, the
flat $K$-step history baseline, and the filler instruments. \textbf{(iii) Findings} M1 to M4 from the first controlled closed-loop comparison of flat,
structured, temporal-graph, distilled, and history-based context (one protocol, three backbones), a
real-data-seeded replay benchmark ($54$ highD cut-ins), and released ports, seeds, caches, and protocols.

\section{Related Work}
\label{sec:related}

\paragraph{Positioning.} No prior driving-memory paper treats the warmup pool as a random factor
(DiLu ablates pool \emph{size}, never composition~\citep{wen2024dilu}), and none measures agent memory
against token-matched irrelevant filler, even though random documents can \emph{improve} RAG
accuracy~\citep{cuconasu2024power} and length alone degrades it~\citep{du2025context}.

\paragraph{Driving-agent memories.}
DiLu~\citep{wen2024dilu} established the closed-loop reason, reflect, and memorize paradigm on
\texttt{HighwayEnv}. LimSim++~\citep{fu2024limsimpp}, LeapAD~\citep{mei2024leapad}, and
SafeDrive~\citep{safedrive2024} extend it, all retrieving over \emph{flat} representations, and other
driving agents with memory or reflection~\citep{fu2023drivelikehuman,mao2024agentdriver} share the same
confound of joint variation. RESPOND~\citep{respond2025}, the closest prior work, retrieves a
$5\!\times\!3$ ego-grid by pattern matching. \method adds the two axes it lacks, temporal structure and
relational retrieval, though our experiments bear out no advantage for either, and our grid port's failed
audit rules out any claim about the published system.

\paragraph{Graph-structured and episodic memory for LLM agents.}
Outside driving, knowledge-graph memories, some of them temporal, report strong recall and context
reductions: Zep~\citep{zep2025}, Mem0 and its graph variant~\citep{mem0_2025}, A-MEM~\citep{axu2025amem},
HippoRAG~\citep{hipporag2024}, and Optimus-1~\citep{optimus1_2024}. That evidence is almost entirely
\emph{open-loop QA}, and mixed even there (Mem0's graph variant trails its flat variant on single-hop and
multi-hop questions). In closed-loop \emph{control} we find only two same-agent comparisons of structured
against flat memory, AriGraph~\citep{arigraph2024} and G-Memory, where flat retrieval baselines often
score \emph{below} no-memory on ALFWorld~\citep{gmemory2025}, and neither is token-matched. The known tendency of unfiltered experience accumulation to degrade
performance~\citep{xiong2025memory} takes a sharper form here: the pool controls the sign.

\paragraph{Curation and reflection.}
That \emph{what} gets stored and injected dominates is anticipated by Reflexion~\citep{shinn2023reflexion},
Voyager~\citep{wang2023voyager}, Generative Agents~\citep{park2023generative}, and DiLu's own curated
pool~\citep{wen2024dilu}. ExpeL~\citep{zhao2024expel} distills insights
open-loop. Our rulebook is its closed-loop instantiation, a distillation control that posts the
evaluation's best collision-only cell while its gains price as caution (\S\ref{sec:consolidated}).

\section{The Probe Suite: Memory Designs and Controls Under Test}
\label{sec:method}

This section details the richest probe, \method, our episodic temporal-graph memory, which we treat
as one point in the design space rather than as the contribution. The other probes are specified where
they are evaluated (\S\ref{sec:setup}, \S\ref{sec:padding}), with full texts in the supplementary material.

\begin{figure}[t]
\begin{center}
\includegraphics[width=0.42\textwidth]{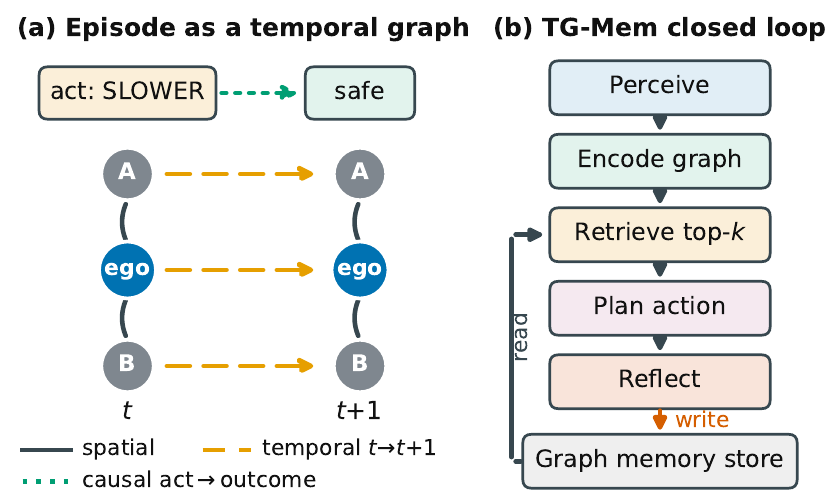}
\caption{\method{} in one view. \textbf{(a)} An episode as a typed temporal graph, with spatial edges
within a frame, temporal $t\!\to\!t{+}1$ edges across frames, and the action and outcome stored as
metadata (model-inferred causal edges are disclosed as such). \textbf{(b)} The perceive $\to$ encode
$\to$ retrieve $\to$ plan $\to$ reflect loop over the graph store.}
\label{fig:method}
\end{center}
\end{figure}

\method maintains a graph-structured episodic memory $\mathcal{M}$ and, at every decision step, executes
a \textbf{perceive $\to$ encode $\to$ retrieve $\to$ plan $\to$ reflect} loop (Figure~\ref{fig:method}).
All steps are LLM-API or graph-algorithm calls, and no parameters are trained.

\paragraph{Episode as a temporal graph.}
An episode $e$ summarizes a $K$-step window and is a typed graph $e=(V,E)$ with a small fixed schema.
\textbf{Nodes} $V$: \texttt{ego}, \texttt{vehicle}$_i$, \texttt{lane}$_j$, \texttt{hazard}$_k$, with
kinematic attributes (relative position and velocity, lane, inferred intent). \textbf{Edges} $E$:
\emph{spatial} (\texttt{ahead\_of}, \texttt{left\_of}, \texttt{merging\_into}, \dots) and \emph{temporal}
(\texttt{next}: entity state at $t\!\rightarrow\!t\!+\!1$). The executed action $a_t$ and outcome
$\omega_t$ are stored as episode \emph{metadata} alongside the graph (Figure~\ref{fig:method}a) and are
rendered into the injected text at retrieval time. The schema also permits \emph{causal} (\texttt{ego\_action}$\rightarrow$\texttt{outcome}) edges, but
the extractor never sees the executed action, so such edges are model-inferred, not grounded. We obtain $e$ by prompting the model with the recent window under a strict JSON schema. A
\emph{programmatic} variant builds the same graph deterministically from the numeric observation, with
exact kinematics and no extraction call.

\paragraph{Encoding, storage, and retrieval.}
The current window is encoded into a query graph $q$ under the same schema, and episodes live in an
in-process store. Retrieval scores every stored episode $e$ against $q$ with a combined
structural--temporal similarity
\begin{equation}
  s(q,e) \;=\; \alpha\, s_{\text{struct}}(q,e) \;+\; \beta\, s_{\text{temp}}(q,e),
  \qquad \alpha+\beta=1,
  \label{eq:score}
\end{equation}
and injects the top-$k$. Both terms are multiset Jaccard overlaps over \emph{typed} elements. The structural term
$s_{\text{struct}}$ averages the overlaps of the node-type bag and the labelled \emph{spatial}-edge bag,
and $s_{\text{temp}}$ is the overlap of the labelled \emph{temporal and causal} edge bag, defaulting to
a neutral $\tfrac12$ when no temporal structure is present. Typed bags make matching \emph{relational}
and time-aware, the intended alternative to flat cosine and single-frame grid lookup. Setting
$\beta\!=\!0$ recovers a static structural matcher (ablated in \S\ref{sec:ablations}).

\paragraph{Injection, planning, and reflection.}
The retrieved episodes are serialized into a compact few-shot block (scene $\rightarrow$ action
$\rightarrow$ consequence) and appended to a shared planning prompt. The planner emits a discrete meta-action via structured
output, and every LLM call's tokens are logged. On a collision, \method
prompts the model for the maneuver that would have avoided the outcome and inserts a \emph{corrected}
episode, retrievable in temporally similar states (identically for every reflecting method).

\section{Experimental Setup}
\label{sec:setup}

\paragraph{Simulator and scenarios.}
We evaluate in closed loop on \texttt{HighwayEnv}~\citep{highwayenv} (v1.11, CPU-only), whose discrete
meta-action space matches LLM decisions, using four scenario families.
\textbf{(1) \texttt{merge-v0}} has multi-step approach, yield, and merge dynamics and is the flagship
synthetic comparison ($n\!=\!100$ evaluation seeds per method under \emph{ten} disjoint $8$-episode
warmup pools plus two $40$-episode pools on the weak backbone, and $n\!=\!10$ on the others).
\textbf{(2) Two constructed diagnostics} (ours): \texttt{stopgo-v0}, a \emph{retrieval-key} diagnostic
(steady vs.\ stop-and-go lead by seed parity, $20{+}20$ episodes, \S\ref{sec:stopgo}), and
\texttt{telegraph-v0}, a \emph{pre-registered history-necessity} diagnostic with a dual success metric
and scripted certificates (\S\ref{sec:telegraph}). \textbf{(3) Urban probes}, \texttt{roundabout-v0} and
\texttt{intersection-v0/v1} (structural nulls). \textbf{(4) \texttt{highd-cutin-v0}}
(ours), reconstructed initial-condition replays of real recorded cut-ins, described below.
Difficulty-calibration probes are in Tables~S24--S29. Warmup and evaluation seed sets are fixed,
disjoint, and released.

\paragraph{Reconstructed initial-condition replays of real cut-ins (\texttt{highd-cutin-v0}).}
We reconstruct the $54$ unique high-risk cut-in events (time-to-collision ${<}4$\,s) bundled with the
RESPOND repository from the highD drone dataset~\citep{highD2018}. These are \emph{initial-condition replays, not full trajectory playback}, because the raw highD
trajectories are license-gated. The ego spawns at its recorded speed and lane, neighbours at their
recorded offsets and speeds, IDM controls the non-ego vehicles from then on, and the LLM drives $15$\,s
in our closed-loop protocol . The replays reproduce each event's initial risk geometry, not its recorded traffic. Trivial policies crash on $50$ to
$78\%$ of them. We warm on $14$ events and evaluate frozen on the remaining $40$.

\paragraph{Memory protocol, and the warmup pool as a random factor.}
For each (method, scenario, backbone) triple we \emph{warm up} over $8$ seeds ($40$ in the budget
experiment, $14$ events for highD) with learning on, then \emph{freeze} the memory and \emph{evaluate}
on the disjoint held-out seed set with retrieval only. No-Memory and the warmup-free arms (history,
padding, fillers) skip warmup. Each warmup pool is a disjoint block of eight independently seeded
simulator episodes. Because the seed universe is effectively unbounded, disjoint blocks are independent
draws from the warmup-generation distribution, which is the population over which the pool-level
estimand, the expected effect of adding memory over $8$-episode warmup draws on this testbed, is defined.

\paragraph{Memory designs compared, and what actually varies.}
All methods share the scene encoder, planning prompt skeleton, and planner. Only the memory system
varies, as a bundle of stored representation, retrieval key, retrieved content, and injection format.
\textbf{No-Memory}
is the zero-shot planner and serves as the control. The \textbf{DiLu port}~\citep{wen2024dilu} embeds
last-frame scene text locally (\texttt{all-MiniLM-L6-v2}), retrieves the cosine top-$k$ over a
\emph{curated} pool, and reflects with the LLM on crashes. The \textbf{grid port}
(RESPOND-inspired)~\citep{respond2025} stores $5\!\times\!3$ ego-grid risk patterns with exact and
bounded sub-pattern matching and LLM reflection. Both are our re-implementations, and every baseline
conclusion below is a statement about these audited ports, bounded by the port audits (Table~S11). The
grid port's planner-substrate and representation audits both fail substantially, so its rows measure
\emph{our port under this harness} and support no conclusion about the published RESPOND system.
The \textbf{flat $K$-step history} baseline (ours) injects the planner's own last $K{-}1$ raw frames
as text, with no storage, retrieval, or reflection. \method{} with its variants, the rulebook, and the
padding and filler arms (all ours) complete the suite under the identical protocol.

\paragraph{Backbones.}
Models are served through OpenRouter~\citep{openrouter} at temperature $0$ with fixed seeds:
\texttt{gpt-4o-mini} (all scenarios and every decisive contrast), \texttt{llama-3.3-70b-instruct}, and
\texttt{mistral-small-3.2-24b}. All calls use strict JSON-schema structured outputs, and every response
is cached by request hash.

\paragraph{Metrics and statistics.}
The primary safety metric is the \emph{collision-free rate}. Because \texttt{merge-v0} terminates when
the ego passes the ramp section, we additionally report caution-pricing metrics recovered by replaying
each logged action sequence through the seeded deterministic environment: mean ego speed,
time-to-completion, timeout share, and a pre-registered composite, \emph{safe-and-swift} (collision-free
\emph{and} no timeout \emph{and} mean speed $\geq 0.9\times$ the No-Memory mean). Scripted non-LLM certificate policies bound what caution alone achieves on each scenario. Every arm
on a scenario is evaluated on the identical seed set, so same-seed contrasts are \emph{paired}. We
report exact McNemar tests on the discordant pairs, Holm-adjusted within each declared family, with the
unpaired Fisher tests of the review version alongside in Table~S50. No headline inference differs
between the two, and the two cells that cross $0.05$ under pairing both strengthen the findings.
Pool-level inference is model-free first. We count the pools in which a memory exceeds the pool-free
No-Memory reference and apply an exact sign test, run per-pool tests with Holm adjustment, and take a
nonparametric bootstrap over the ten pool-level effects for the pool-averaged effect. A
random-intercept binomial model is retained as a supplementary effect-size estimate (Table~S45).
Per-decision calls, tokens, and cost count \emph{every} call, including \method's extraction. Cells with
$n\!\leq\!20$ per condition are \emph{exploratory probes} and are labeled as such where they appear.
Confirmatory claims rest on the $n\!=\!100$ merge and $n\!=\!40$ diagnostic cells. Hypotheses fixed
before execution are marked \emph{pre-registered} (full text released in \texttt{PREREGISTRATION.md}).
Everything else is exploratory or post-hoc and labeled in place.

\section{Results}
\label{sec:results}

\subsection{Synthetic multi-step merges}
Table~\ref{tab:main} reports \texttt{merge-v0} at $n\!=\!100$ evaluation seeds per method, with memory
warmed on pool A and then frozen, on \texttt{gpt-4o-mini}. All arms share the seed set, so every
contrast is a paired exact McNemar test, Holm-adjusted within the four warmup arms (the full table with
the unpaired Fisher tests alongside is Table~S50). Four facts stand out.
\textbf{(i)} Both well-designed memories sit about $0.15$ above No-Memory: the DiLu port at $0.670$
[$0.57,0.75$] and \method at $0.660$ [$0.56,0.75$] against $0.510$ [$0.41,0.61$] (Wilson $95\%$
intervals). Under pairing the \method{} margin is Holm-robust ($24$ vs.\ $9$ discordant seeds,
$p_{\text{holm}}\!=\!0.041$) and the DiLu port's is marginal ($34$ vs.\ $18$, $p_{\text{holm}}\!=\!0.073$).
Neither survived Holm under the unpaired test. The margin is real \emph{for this pool} and, as shown below, a property of the draw. \textbf{(ii)} The DiLu port and \method are indistinguishable ($17$ vs.\ $16$ discordant seeds,
$p\!=\!1.0$), while \method pays twice the calls and four times the tokens ($1{,}038$ prompt tokens per
decision against $210$ to $274$). Its extra computation buys no safety, and doubling the retrieved
context \emph{reduces} safety (Figure~S1). \textbf{(iii)} The flat $K$-step
history baseline, which hands the planner its own last four frames with no memory system, scores
$0.750$. That is above No-Memory ($p\!=\!2\times10^{-5}$), not separable from the DiLu port ($p\!=\!0.30$),
and inside the filler band (vs.\ $0.800$, $p\!=\!0.47$). Real temporal context helps merge about as much as irrelevant filler does. \textbf{(iv)} The grid port is significantly below \emph{everything}, including
No-Memory ($p_{\text{holm}}\!\leq\!3\times10^{-9}$). Its rows measure \emph{our port under this harness},
not the published RESPOND system (the audit it fails is Table~S11). Under our earlier crippled ports the DiLu port scored roughly \emph{half} its current value.

\begin{table}[t]
\centering
\caption{Closed-loop \texttt{merge-v0} ($n\!=\!100$/method, one shared seed set, \texttt{gpt-4o-mini},
audited v2 ports, warmup pool A). Paired McNemar tests, Holm-adjusted over the four warmup arms: grid
port below everything ($p_{\text{holm}}\!\leq\!3\times10^{-9}$), \method{} vs.\ No-Memory
$p_{\text{holm}}\!=\!0.041$, DiLu port vs.\ No-Memory $0.073$, DiLu port vs.\ \method{} $1.0$. \emph{Pool A
is the best of ten draws, and under B--J the memory rows collapse to $0.08$--$0.48$} (Figure~\ref{fig:pools}).
SaS is the pre-registered safe-and-swift composite (\S\ref{sec:setup}), and speed is in m/s from certified
replays. History and padding rows are warmup-free controls outside the Holm family (raw paired $p$:
padding above \method{} $0.029$, above DiLu port $0.024$). Lower block: \method{} ablation on the same pool
($n\!=\!40$ shared seeds, full \method{} $0.700$, paired McNemar vs.\ full and unpaired Fisher,
$^{\ast}p\!<\!0.05$, pool-conditional, \S\ref{sec:ablations}). Tokens count \emph{every} LLM call.}
\label{tab:main}
\small
\setlength{\tabcolsep}{4.5pt}
\begin{tabular}{lccccc}
\toprule
Method & Coll.-free $\uparrow$ & SaS $\uparrow$ & Speed & Steps & Tok./dec.\ $\downarrow$ \\
\midrule
No-Memory           & 0.510 & 0.40 & 24.2 & 12.7 & 210 \\
DiLu port           & 0.670 & \textbf{0.67} & 25.6 & 12.1 & 274 \\
Grid port (RESPOND-inspired) & 0.140 & 0.14 & 29.0 & \phantom{0}7.4 & 251 \\
\method{} (ours)    & 0.660 & 0.45 & 23.2 & 13.6 & 1{,}038 \\
\midrule
Flat $K$-step history (ours) & 0.750 & 0.31 & 22.5 & 14.7 & 527 \\
\emph{Irrelevant-padding control (ours)} & \textbf{0.800} & 0.29 & 22.3 & 15.4 & 1{,}037 \\
\midrule
Ablation ($n{=}40$): & Coll.-free & \multicolumn{4}{c}{McNemar $p$ / Fisher $p$} \\
\quad \method{} $-$ temporal edges & 0.550 & \multicolumn{4}{c}{$0.070$ / $0.25$} \\
\quad \method{} $-$ reflection & 0.475 & \multicolumn{4}{c}{$0.004^{\ast}$ / $0.056$} \\
\quad \method{} single-frame window ($K{=}1$) & 0.400 & \multicolumn{4}{c}{$0.017^{\ast}$ / $0.007^{\ast}$} \\
\bottomrule
\end{tabular}
\end{table}

\paragraph{Scripted certificates bracket the task.} Three non-LLM policies ($n\!=\!100$) calibrate the
collision-only metric: always-\texttt{IDLE} scores $0.00$, a ten-line reactive gap-keeper $0.40$, and
always-\texttt{SLOWER} $\mathbf{0.98}$ at $21.0$\,m/s. The ceiling is reachable by blanket caution alone,
so we read every number against it and price caution explicitly (the SaS column).

\begin{figure}[t]
\begin{center}
\includegraphics[width=0.62\textwidth]{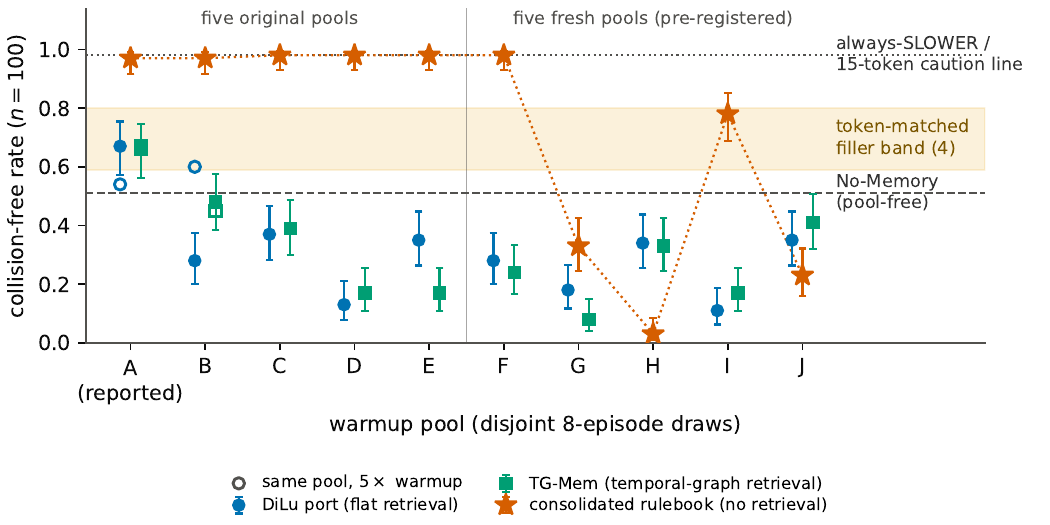}
\caption{The warmup-pool lottery on \texttt{merge-v0} over all ten disjoint $8$-episode pools
($n\!=\!100$ per cell, Wilson $95\%$ intervals) against the pool-free No-Memory arm (dashed), the band
spanned by the four token-matched fillers, and the always-\texttt{SLOWER} and caution-line ceiling
(dotted). Both retrieval memories exceed No-Memory only under pool A. Open markers re-warm pools A and B
on five times the budget. The retrieval-free rulebook (stars) looked pool-immune on the five original
pools and fell to $0.03$--$0.78$ on three of the five pre-registered fresh pools.}
\label{fig:pools}
\end{center}
\end{figure}

\paragraph{The margin is the best of ten draws.} Every merge cell above shares warmup pool A. Treating
the pool as the random factor it is, we re-ran both memories under nine more disjoint $8$-episode pools
(B to J, $n\!=\!100$ each, with No-Memory, which takes no warmup, as the shared comparator at
$0.510$). Figure~\ref{fig:pools} plots every cell and Tables~S45 and S52 list them. The DiLu port ranges
from $0.110$ to $0.370$ and \method{} from $0.080$ to $0.480$ over the nine new pools, so pool A is the
\emph{best of ten draws for both methods}. The model-free
statement is the strongest one. Each memory exceeds No-Memory in exactly one pool of ten, which under the
null that the median pool effect of adding memory is non-negative has probability $11/1024\!=\!0.011$
(exact sign test, per method), and in five of the other nine pools each memory is significantly
\emph{below} No-Memory after Holm adjustment (Table~S45). A nonparametric bootstrap over the ten
pool-level effects places the pool-averaged effect of adding memory at $-0.20$ [$-0.29,-0.10$] for the
DiLu port and $-0.20$ [$-0.30,-0.09$] for \method{} (Table~S51). The sign is secure, the magnitude
coarse. The pre-registered hierarchical random-intercept model agrees ($-0.22$ [$-0.32,-0.12$] for both,
Table~S45) and now serves as a supplementary effect-size estimate rather than the primary evidence. Re-warming on five times the budget (pools A40 and B40) \emph{stabilizes the
lottery} (DiLu port $0.540/0.600$ from $0.670/0.280$) but to \emph{no benefit}: no $40$-episode cell
differs from No-Memory after Holm adjustment (open markers in Figure~\ref{fig:pools}).

\paragraph{Our token-matched irrelevant-padding control.}\label{sec:padding} Memories inject more prompt
than No-Memory ($274$ to $1{,}038$ vs.\ $210$ tokens per decision), which confounds content with
length. Our control injects fixed, explicitly irrelevant, non-advisory filler (Table~S43) matching
\method's \emph{total} footprint ($1{,}037$ tokens per decision) with no retrieval, storage, or
reflection. On merge the control scores $\mathbf{0.800}$: above No-Memory (paired
$p\!=\!2\times10^{-6}$), above \method ($p\!=\!0.029$), and above the DiLu port ($p\!=\!0.024$, a cell the
unpaired test had left marginal). \emph{No memory tested exceeds any filler's point estimate} (the weaker fillers, $0.59$ to $0.73$,
are within noise of the memories rather than demonstrably equivalent), and the spread across arbitrary
fillers exceeds the claimed memory margin ($+0.15$), so merge safety varies more with prompt composition
than with any memory design tested. A pre-registered disclaimer$\times$domain factorial (Table~S46) shows the filler factors interact:
removing the ``safe to ignore'' disclaimer costs $0.05$ in one content family ($p\!=\!0.50$) and $0.24$
in the other ($p\!=\!3\times10^{-4}$). Sharper still is the \emph{caution-line} arm, a single $15$-token
imperative sentence (``drive defensively; when uncertain, prefer \texttt{SLOWER}''). It scores
$\mathbf{0.980}$, level with the rulebook ($p\!=\!1.0$) and the always-\texttt{SLOWER} certificate, and
\emph{above} every filler of about $1{,}000$ tokens ($p\!\leq\!5\times10^{-5}$) and every memory
($p\!\leq\!4\times10^{-9}$). The planner simply obeys it ($99.9\%$ \texttt{SLOWER} actions). Padding shifts early actions toward \texttt{SLOWER} ($33\%\!\to\!45\%$,
Figure~\ref{fig:overview}d), that is, earlier braking, which merge rewards and the diagnostic punishes.
At the trajectory level the ceiling arms never fall below a $2$\,s time-to-collision, while the DiLu
port's pool-A gain comes with the closest following of any non-crashing arm (Table~S53). A fourth filler matched to the planner-\emph{visible} footprint ($320$ tokens) scores $0.730$ on
merge and $\mathbf{0.80}$ on the diagnostic ($n\!=\!20$, exploratory). There the graph variants sit far
\emph{below} it ($p\!<\!10^{-4}$, content harm), and the DiLu port's $0.95$, the study's largest memory
effect, is \emph{not separable} from it ($p\!=\!0.34$), so caution-shifting composition may carry most
of that effect.

\paragraph{The consolidated rulebook: pool-immunity on five pools, falsified by five more.}\label{sec:consolidated}
If retrieval amplifies pool noise, replacing it should shrink the lottery. Our pre-registered
\emph{consolidated rulebook} distills the identically gathered pool \emph{once} into at most eight
prescriptive rules (in the style of ExpeL~\citep{zhao2024expel}) and injects the same rulebook at every
decision, with no per-decision key. On the five original pools it posts the study's best learned cells,
$0.97$ to $0.98$ (a range of $0.01$ against the DiLu port's $0.54$), above the strongest filler
($p{=}2\times10^{-4}$ per pool, Figure~\ref{fig:pools} and Table~S52). \emph{Five fresh pools falsify that
reading.} Under F to J, registered with the falsifier ``any pool below the strongest filler,'' the same
procedure scores $0.98$, $0.33$, $0.03$, $0.78$, and $0.23$. Three pools fall below every filler
($p\!\leq\!2\times10^{-11}$) and below No-Memory. Across all ten pools the rulebook's early-\texttt{SLOWER}
share orders its collision-free rate almost perfectly (Spearman $\rho\!=\!0.89$, Table~S52).
Consolidation de-noises \emph{which} lessons reach the planner, but which \emph{disposition} gets
distilled is itself the lottery. The high-scoring pools shift $78$ to $100\%$ of first-five-step actions to \texttt{SLOWER} (filler
$45\%$, baseline $33\%$), and the pre-registered safe-and-swift composite scores them $0.00$
(Table~\ref{tab:main}, Table~S47). Their operating point ($21.0$\,m/s over $16.9$ steps) is behaviorally identical to
always-\texttt{SLOWER}'s and falls below the pre-registered progress floor, while the retrieval arms
keep most of their score (DiLu port $0.67$, \method{} $0.45$, No-Memory $0.40$). Pricing progress even
mildly erases the merge headline. The diagnostics agree (\S\ref{sec:telegraph}). Lessons do not stack with composition either: the DiLu port padded to the visible-matched dose
collapses to $0.26$ to $0.30$ on pools A, B, and D, below the port alone \emph{and} filler alone
(pre-registered, Table~S48).

\paragraph{Model dependence, and structural nulls.} On \texttt{Llama-3.3-70B}, \texttt{merge-v0} is
nearly saturated (every method at or above $90\%$ collision-free, Figure~S2), and a medium-density merge
ties all memories ($n\!=\!10$, exploratory, Table~S6). On the real cut-ins \method is numerically best
with the strong backbone ($0.891$ vs.\ $0.783$, $n\!=\!46$, Holm $p\!=\!1.0$, Table~S10), in an evaluation
that was enlarged \emph{after} we observed the lead. 

\paragraph{Decision frequency: a boundary condition (exploratory).} Everything above runs at this
literature's universal $1$\,Hz. An exploratory sweep ($n{=}20$ per cell, Table~S49) finds the memory-vs-none margin \emph{monotone
decreasing} in decision frequency (DiLu port $+0.40$ at $0.5$\,Hz, $+0.16$ at $1$\,Hz, $-0.30$ at
$3$\,Hz), crossing zero near that convention. The sign change bounds every conclusion above to the
cadence at which it was measured.

\subsection{A retrieval-key diagnostic: stop-and-go leads}
\label{sec:stopgo}
\texttt{stopgo-v0} was built to favor temporal memory. The ego follows a lead that is either steady or a
hard stop-and-go oscillator, identical \emph{at spawn}, so identifying the lead requires history. The
result inverted our expectation. On stop-and-go episodes ($n\!=\!20$ per cell, exploratory scale
throughout this subsection), curated flat memory reaches $\mathbf{0.95}$ versus $0.45$ for No-Memory
($p\!=\!0.001$), while \method drops to $0.10$, \emph{significantly worse than no memory} ($p\!=\!0.03$),
and the single-frame grid to $0.00$. The flat $K$-step history baseline scores $0.65$ (n.s., with $1.00$
in the steady regime). \textbf{What the diagnostic measures.} Once braking begins the closing speed is
visible in the frame, and the DiLu port's key is exactly that frame's text, so its reflected slow-down
lessons fire when the frame matches. The diagnostic does not show that temporal keys are necessary. It shows that \emph{key precision} decides: precise single-frame keys ($0.95$) beat coarse type-bag
keys ($0.00$ to $0.10$). \textbf{Brittle.} Stress-tested (Tables~S32--S35), the DiLu port scores $0.95$ and $0.80$ under two
warmup splits but $0.10$, $0.05$, and $0.00$ under a third split, a faster oscillation, and a larger
gap: sign instability rather than reliable benefit. \textbf{Second and third backbones.} On
\texttt{Llama-3.3-70B} (Table~S40) No-Memory solves the stop-and-go regime $20/20$, so every separation
above is a weak-backbone phenomenon, and \method drags the solved regime to $0.60$
($p_{\text{holm}}\!=\!0.033$), but so does the padding control. On \texttt{Mistral-Small-24B}
(Table~S42) neither memory moves the partially saturated task (No-Memory $0.80$, both memories $0.75$,
filler $0.95$). Nowhere does memory beat No-Memory or filler.

\paragraph{Curation vs.\ key precision: a double dissociation.} Two further DiLu-port cells,
uncurated-semantic and curated-lexical keys, dissociate the ingredients. On \texttt{merge-v0} (pool A, $n\!=\!100$), removing curation collapses the port to $0.490$
($p\!=\!0.015$) while lexical keys hold $0.650$: curation necessary, precision not. On the diagnostic,
under all three warmup splits (Tables~S23 and S38--S39), uncurated-semantic scores $1.00/1.00/0.70$ and
curated-lexical $0.00/0.00/0.15$ ($p\!<\!0.002$ per split): precision necessary, curation not.

\paragraph{Post-mortem and the completed $2\!\times\!2$ (supplementary \S G).} A six-step elimination
chain (Tables~S12--S13 and S19--S23) fails to rescue \method{} on the diagnostic. The completed key-type
$\times$ injection-format $2\!\times\!2$ (Table~S44, $n\!=\!20$ per cell) finds one robust culprit,
compact injection, which is harmful under every warmup split, while the key-type attribution flips with
the pool. Even mechanism claims inherit the lottery.

\subsection{A pre-registered history-necessity diagnostic: telegraphed slams}
\label{sec:telegraph}
Its pre-registered successor removes that leak and closes the caution-dial loophole. In \texttt{telegraph-v0} the ego follows a lead that either
\emph{telegraphs} (two decelerate-and-recover taps at the step-4 and step-6 observations, inside the
$K{=}5$ window, then a hard slam just after step 8) or is benign. Taps are position-compensated and the
spawn gap is jittered by $\pm2$\,m, so pre-slam text states are indistinguishable across types, and the
lead's future braking is predictable \emph{only} from its past
signature~\citep{wang2018braking,sunberg2022internal}. Success requires \textbf{both} no collision
\textbf{and} a progress target that blanket caution cannot reach. Scripted certificates fix well-posedness before any LLM run, and a type-aware oracle passes both
regimes ($1.00/1.00$).

No design tested passes the pre-registered criteria. Every arm run (No-Memory, DiLu port, programmatic
\method, both fillers, Table~S30, plus the rulebook and history arms) scores $0/20$ on hazard episodes
under both warmup pools, and the failure \emph{modes} are the finding. Under pool A, the DiLu port is the
only cell that crashes, an \emph{aggressive} disposition with a perfect benign score, while every other
memory survives the slam but times out. Under pool B the same method flips wholesale to \emph{timid},
with benign success going from $1.00$ to $0.00$ and the failure mode from all-crash to all-timeout. The
lottery controls a disposition, not a score (Figure~\ref{fig:overview}c). \method's temporal keys, which
by construction \emph{can} separate the regimes, do not rescue it. Neither does removing the memory system: the \emph{flat $K$-step history} baseline, with both taps
in its window at slam time, survives every hazard episode ($20/20$) yet reaches the progress target on
$0/20$, and on only $9/20$ benign episodes, buying indiscriminate caution rather than the selective
slow-then-resume behavior the oracle certifies as achievable. This arm also bounds an oracle-retrieval experiment: perfect access
to the needed frames does not produce the certified behavior, so the bottleneck here is not retrieval
precision alone. The rulebook never crashes and never progresses ($40/40$ timeouts under both pools).

\subsection{Reconstructed initial-condition replays of real cut-ins (highD)}
\looseness=-1 These replays reconstruct each event's initial risk geometry with IDM-controlled
neighbours, not full recorded trajectories (\S\ref{sec:setup}), so they do not establish robustness in
realistic closed-loop driving. On the 40 held-out real cut-ins
(\texttt{gpt-4o-mini}, memories warmed on 14 disjoint events), the no-memory agent is collision-free on
$0.875$ of episodes and \emph{no} memory improves on it (\method $0.775$, DiLu port $0.800$, grid port
$0.825$, all n.s., $p\!\geq\!0.37$, Table~S8). A nearly saturated baseline leaves little headroom, so this
null is weak evidence either way. It completes the map on this testbed: \emph{memory helped only where
difficulty matched capability, and only under pools whose lessons happened to transfer.}

\subsection{Ablations}
\label{sec:ablations}
We isolate each component of \method on \texttt{merge-v0} (pool A, $n\!=\!40$, lower block of
Table~\ref{tab:main} and Table~S36). Removing reflection ($0.475$) is significant under pairing
($p\!=\!0.004$), the single-frame window ($0.400$) under both tests, and removing temporal edges
($0.550$) marginal at best ($p\!=\!0.070$). The window and reflection matter, while the temporal edges, the central novelty, contribute at
most marginally, and since the ablation ran on pool A alone, this marginal effect is pool-conditional
and, if anything, weaker evidence for temporal edges. Hyperparameter sweeps agree (Figure~S1, $n\!=\!10$):
every off-default $(K,k)$ configuration degrades.

\section{Limitations}
\label{sec:limitations}

\paragraph{Generalizability.} Every decisive contrast comes from \texttt{HighwayEnv} with kinematic
observations, discrete meta-actions, temperature-$0$ decoding, the literature's $1$\,Hz cadence, warmup
pools of $8$ to $40$ episodes, and \texttt{gpt-4o-mini}. The findings characterize a weak-planner regime in a simplified control environment and do not
establish that pool composition or prompt effects dominate memory in richer environments or other
closed-loop domains. The frequency sweep (Table~S49) is an explicit boundary condition, since
the memory-vs-none margin changes sign between $0.5$ and $3$\,Hz, so every conclusion is conditional on
the cadence. The highD experiments are reconstructed initial-condition replays on a nearly saturated
baseline, not realistic closed-loop driving. Richer simulators (CARLA, nuPlan) are the next step.

\paragraph{Statistics.} Ten pools estimate the pool-averaged effect coarsely, so the sign test and
the bootstrap support the sign of the average effect, not its magnitude. Where we say a filler is not beaten, we report point-estimate orderings with intervals rather
than equivalence tests.

\paragraph{Filler mechanism, method space, and ports.} The filler factorial, dose response, and caution-line arm only partially separate the filler
channels, since length, position, style, and instruction salience remain incompletely crossed. ``Caution dial'' names the dominant
behavioral signature we measure (the early-\texttt{SLOWER} shift), not an identified mechanism. Larger
continual regimes, learned retrieval, and oracle-retrieval upper bounds are untested, and the ablation
weighing \method's temporal edges ran on pool A, so its attribution is pool-conditional. Baseline
conclusions concern our audited ports under this harness (Table~S11), not the published systems. The
grid port is a best-effort adaptation of a partially released system and says nothing about the
published RESPOND. 

\section{Conclusion}
\label{sec:conclusion}

We asked when experience memory helps closed-loop LLM driving agents and answered for one audited
testbed with a controlled evaluation built on resampled warmup pools, token-matched fillers,
oracle-certified diagnostics, caution-pricing metrics, and a probe suite spanning flat, structured,
temporal-graph, distilled, and raw-history designs. On this testbed, memory effects were dominated by
pool and prompt composition rather than by what the memory stored, and even our own strongest claim
survived only until the evaluation's resampling control reached it. These are statements about
\texttt{HighwayEnv} at $1$\,Hz with a weak planner, not about closed-loop LLM driving in general. The
next experiments are an \emph{oracle-retrieval} arm, which separates retrieval failure from the
planner's inability to use correct retrievals, and a port of the control stack to CARLA or nuPlan.
Future evaluations should resample warmup pools, add token-matched filler controls, validate ports,
pair same-seed comparisons, report family-corrected and pool-level statistics, match difficulty to
capability, and price caution in the metric.

%\acks{TODO (optional): funding and thanks. Uncomment if used; this counts toward the 16 pages.}

\bibliography{references}

\end{document}